\documentclass{ecai}
\usepackage{times}
\usepackage{graphicx}
\usepackage{latexsym}

\usepackage{array}
\usepackage{latexsym}
\usepackage{bbm}
\usepackage{xcolor}
\usepackage{array}
\usepackage{tabularx}
\usepackage{amsmath}
\usepackage{amssymb}
\usepackage{bm}
\usepackage{booktabs,dcolumn}
\usepackage{url}
\usepackage{caption}
\usepackage[linesnumbered,ruled]{algorithm2e}
\usepackage{multirow}

\newcommand{\ra}[1]{\renewcommand{\arraystretch}{#1}}
\newcolumntype{?}{!{\vrule width 1pt}}

\begin{document}

\title{DAN: Dual-View Representation Learning for Adapting Stance Classifiers to New Domains}

\author{Chang Xu$^1$ \and C\'ecile Paris$^1$ \and Surya Nepal$^1$ \and Ross Sparks\institute{CSIRO Data61, Australia, email: \{Chang.Xu, Cecile.Paris, Surya.Nepal, Ross.Sparks\}@data61.csiro.au}\\
 Chong Long$^2$ \and Yafang Wang\institute{Ant Financial Services Group, email: \{huangxuan.lc, yafang.wyf\}@antfin.com}}

\maketitle
\bibliographystyle{ecai}

\begin{abstract}
We address the issue of having a limited number of annotations for stance classification in a new domain, by adapting out-of-domain classifiers with domain adaptation.
Existing approaches often align different domains in a single, global feature space (or view), which may fail to fully capture the richness of the languages used for expressing stances, leading to reduced adaptability on stance data.
In this paper, we identify two major types of stance expressions that are linguistically distinct, and we propose a tailored \textbf{d}ual-view \textbf{a}daptation \textbf{n}etwork (DAN) to adapt these expressions across domains.
The proposed model first learns a separate view for domain transfer in each expression channel and then selects the best adapted parts of both views for optimal transfer.
We find that the learned view features can be more easily aligned and more stance-discriminative in either or both views, leading to more transferable overall features after combining the views.
Results from extensive experiments show that our method can enhance the state-of-the-art single-view methods in matching stance data across different domains, and that it consistently improves those methods on various adaptation tasks.
\end{abstract}

\section{Introduction}
There has been a growing interest in the relatively new task of stance classification in opinion mining, which aims at automatically recognising one's attitude or position (e.g., \textit{favour} or \textit{against}) towards a given controversial topic (e.g., feminist movement)~\cite{walker2012stance,hasan2013stance,mohammad2016semeval,bar2017stance}.
Recently, deep neural networks (DNNs) have been used to learn representations for stance expressions, resulting in state-of-the-art performance on multiple stance corpora~\cite{augenstein2016stance,li2017deep,du2017stance,sun2018stance}.
However, DNNs are notorious for relying on abundant labelled data for training, which could be hardly available for a new trending topic, as obtaining quality stance annotations is often costly~\cite{mohammad2017stance}.

To address this issue, domain adaptation~\cite{blitzer2006domain} enables adapting what has been learned from a \textit{source} domain to a \textit{target} domain, usually by aligning the source and target data distributions in a shared feature space.
This process makes the learned features invariant to the domain shift and thus become generalisable across the domains.
Recently, due to their effectiveness and seamless integration with DNNs, \textit{adversarial} adaptation methods~\cite{ganin2015unsupervised,shen2018wasserstein} have gained popularity among various NLP tasks~\cite{kim2017adversarial,zhang2017aspect,alam2018domain,li2018hierarchical}.
In these approaches, a \textit{domain examiner} (also called \textit{domain classifier}~\cite{ganin2015unsupervised} or \textit{domain critic}~\cite{arjovsky2017wasserstein}) is introduced to assess the discrepancy between the domains, and, by confusing it with an adversarial loss, one obtains domain-invariant features.

However, as the domain examiner solely induces a global feature space (view) for aligning the domains, it might not fully capture the various ways stances are expressed in real-world scenarios.
For example, Table~\ref{tb:stance_variation} shows examples of commonly observed stance-bearing utterances where stances are expressed in two distinct ways: \textit{explicitly} via \textit{subjective} expressions carrying opinions (e.g., ``really incredible'') and/or \textit{implicitly} via \textit{objective} expressions that provide facts to support the stance.
With only a single feature space, the distributions of different expression types could be intertwined in the space, which could hinder the domains from being optimally aligned, leading to inferior adaptation performance.

\begin{table}
	\ra{1.1}
	\caption{Examples of stances conveyed explicitly (with opinions) or implicitly (with facts).}
	\footnotesize
	\begin{tabular}{|p{.55\columnwidth}|p{.17\columnwidth}|c|}
		\hline
		Utterance & Topic & Stance\\
		\hline
		(1) Its really incredible how much this world needs the work of missionaries! (\textbf{Opinion}) & Atheism  & Against \\ 
		\hline
		(2) Women who aborted were 154\% more likely to commit suicide than women who carried to term. (\textbf{Fact}) & Legalisation of Abortion  &  Against \\
		\hline
		\multicolumn{3}{l}{\textit{Source: SemEval-2016 Task 6}}\\
	\end{tabular} 
	\label{tb:stance_variation}
\end{table}

In this paper, to cope with the heterogeneity in stance-expressing languages and adapt stance classifiers to the shift of domains, we first identify the aforementioned types of stance expressions, i.e., \textit{subjective} and \textit{objective} stance expressions, as the major elements for a better characterisation of a stance-bearing utterance.
In particular, we propose a hypothesised \textit{dual-view stance model} which regards a stance-bearing utterance as a \textit{mixture} of the subjective and objective stance expressions.
Under this hypothesis, the characterisation of a stance-bearing utterance is then reduced to modelling the more fine-grained subjective and/or objective expressions appearing in the utterance, each of which can receive a different, finer treatment.
Moreover, to implement this dual-view stance model, we propose DAN, the \textbf{d}ual-view \textbf{a}daptation \textbf{n}etwork, for adapting stance classifiers with signals from both the subjective and objective stance expressions.
Specifically, DAN aims at learning transferable subjective and objective features that are both \textbf{domain-invariant and stance-discriminative in their respective views (i.e., feature spaces)}.
To achieve this, DAN is designed to perform three view-centric subtasks: 1) \textit{view formation} that creates the subjective/objective views for learning the view-specific, stance-discriminative features; 2) \textit{view adaptation} that employs a view-specific domain examiner to make each of the view features domain-invariant; and 3) \textit{view fusion} where the view features are made more transferable after being fused in an optimal manner.
All these subtasks are trained jointly in DAN with standard back-propagation.
\begin{figure}[t]
	\centering
	\begin{tabular}{@{}c@{}}
		\includegraphics[width=0.48\columnwidth]{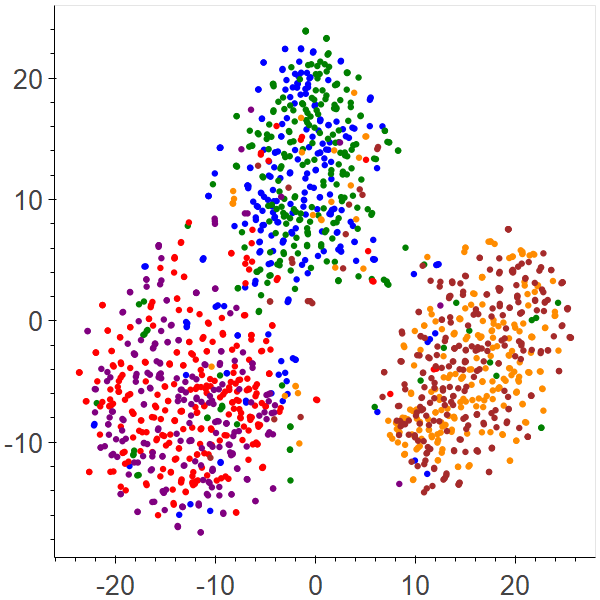}\\
		\scriptsize (a) A single-view method (DANN)
		\label{fig:single}
	\end{tabular}
	~
	\begin{tabular}{@{}c@{}}
		\label{fig:fused}
		\includegraphics[width=0.48\columnwidth]{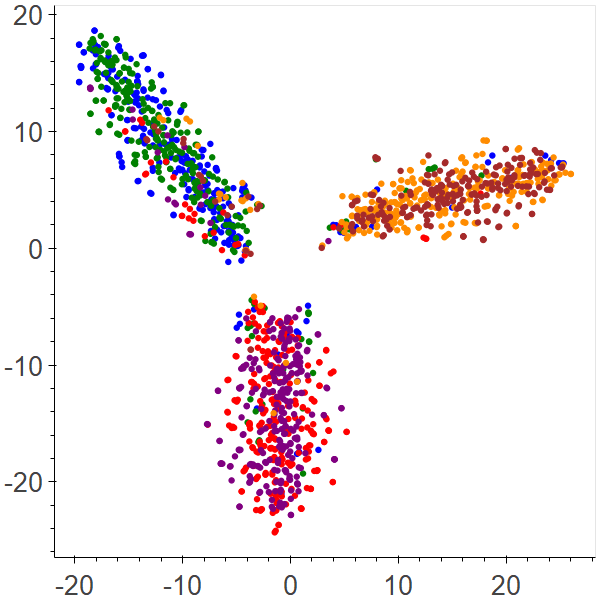}\\
		\scriptsize (b) Our method
	\end{tabular}
	\vspace{-4mm}
	\caption{Feature spaces learned by a state-of-the-art single-view method (a) and our method (b) on a common adaptation task. 
		Source samples are coloured by orange (\textit{favour}), blue (\textit{against}), and red (\textit{neutral}), while target samples by brown (\textit{favour}), green (\textit{against}), and purple (\textit{neutral}).
		The features produced by our method are more domain-invariant and stance-discriminative than those by the single-view method.
		The feature dimension is reduced to 2 by using t-SNE for better visualisation.
	}
	\vspace{-4mm}
	\label{fig:dual_view_illustration}
\end{figure}

We evaluate our method extensively via both quantitative and qualitative analyses on various adaptation tasks.
We find that DAN can enhance the single-view adaptation methods by delivering more transferable features.
As an example, Figure~\ref{fig:dual_view_illustration} shows the features learned by a state-of-the-art single-view approach (DANN~\cite{ganin2016domain}) and its DAN-enhanced version (our method) on a common adaptation task.
As shown, DANN can produce features with good but limited transferability (Figure~\ref{fig:dual_view_illustration}.a): in terms of domain invariance, features of the source and target samples are aligned, but they are scattered around a relatively wide area, indicating non-trivial distances between the source and target features; in terms of stance discrimination, samples from different classes are separated, but the boundary between the \textit{against} and \textit{neutral} classes is still not fully clear.
In contrast, after being enhanced by the proposed DAN to separately adapt the subjective and objective views, the learned features (after view fusion) in the enhanced feature space of DAN (Figure~\ref{fig:dual_view_illustration}.b) exhibit much better transferability: not only do the source/target features become more concentrated, but they are also well separated over the stance classes\footnote{We also visualise the features in the intermediate subjective and objective views of DAN in our experiments (Figure~\ref{tbl:feature_space}).}.
This result suggests that our dual-view treatment of the stance-bearing utterances can yield a more fine-grained adaptation strategy that enables better characterising and transferring heterogeneous stance expressions.

\section{Dissecting Stance Expressions}
To better characterise stance-bearing utterances for domain transfer, we identify two major types of stance expressions to facilitate fine-grained modelling of stance-bearing utterances, which are the opinion-carrying \textit{subjective} stance expressions and the fact-stating \textit{objective} stance expressions.

\textbf{Subjective stance expressions:}
This type of expressions is common in stance-bearing utterances. 
When stating a stance, people may also express certain feelings, sentiments, or beliefs towards the discussed topic, which are the various forms of \textit{subjective} expressions~\cite{wiebe2000learning}.
The role of sentiment information in stance classification has recently been examined~\cite{sobhani2016detecting,mohammad2017stance}, and the identification of sentiment-bearing words in an utterance has been shown to help recognise its stance.
For instance, the underlined sentiment words in the following utterances reveal various stances towards the topic of \textit{Feminist Movement},
\begin{itemize}
	\footnotesize 
	\item[-] \textit{Women are \underline{strong}, women are \underline{smart}, women are \underline{bold}}. (\textbf{Favour})
	\item[-] \textit{My feminist heart is so \underline{angry} right now, wish I could scream my \underline{hate} for inequality right now}. (\textbf{Favour})
	\item[-] \textit{The concept of \#RapeCulture is a \underline{puerile}, \underline{intellectually dishonest} glorification of a crime by feminists}. (\textbf{Against})
\end{itemize}

Based on this observation, we seek to find such stance-indicative subjective expressions in an utterance for stance determination.

\textbf{Objective stance expressions:}
A stance can also be expressed objectively, usually by presenting some facts for backing up the stance.
For example, all the following utterances mention particular evidence for supporting their stances towards \textit{Legalisation of Abortion},
\begin{itemize}
	\footnotesize 
	\item[-] \textit{Life Fact: At just 9 weeks into a pregnancy, a baby begins to sigh and yawn}. (\textbf{Against})
	\item [-] \textit{There are 3000 babies killed DAILY by abortion in the USA}. (\textbf{Against})
	\item[-] \textit{Over the last 25 years, more than 50 countries have changed their laws to allow for greater access to abortion}. (\textbf{Favour})
\end{itemize}

In such case, no explicit subjective signals (e.g., sentiment or emotional words) can be observed; a stance is instead implied in text providing some facts related to the stance.
Usually, such factual information would serve as the reasons for supporting the stances~\cite{hasan2014you,boltuvzic2014back}, thus it may also be stance-specific and become stance-indicative.
Therefore, a different treatment from the one for characterising the subjectivity is needed for capturing such (implicit) objectivity.

\textbf{A dual-view stance model:}
Motivated by the observations made above, we propose a hypothesised \textit{dual-view stance model} to characterise the subjective and objective aspects of stance-bearing utterances, aiming at learning more transferable representations.
Specifically, in this model, we regard any stance-bearing utterance as \textit{a mixture of the subjective and objective stance expressions}.
Formally, given a stance-bearing utterance $\textbf{x}$, we use the following transformation $U$ for such dual-view characterisation,
\begin{equation}
\label{eq:dual_view_model}
\textbf{f}_{\text{dual}}=U(F_{\text{subj}}(\textbf{x}), F_{\text{obj}}(\textbf{x});\theta_U)
\end{equation}
where $F_{\text{subj}}$ and $F_{\text{obj}}$ are the two \textit{view feature functions} (or \textit{view functions} for short) for extracting the subjective and objective features of $\textbf{x}$, respectively.
$\textbf{f}_{\text{dual}}$ is the \textit{dual-view stance feature} of $\textbf{x}$, resulting from applying $U$ to unify both view features provided by the view functions $F_{\text{subj}}$ and $F_{\text{obj}}$.
$\theta_U$ denotes the parameters of $U$, characterising how much contribution from each view to the overall adaptation.
Based on this dual-view stance model, we formulate \textbf{our task of dual-view adaptation of stance classifiers} as follows\footnote{We confine ourselves to the \textit{unsupervised} domain adaptation setting in this work, where we lack annotations in the target domain.}:
given a set of \textit{labelled} samples $S=\{(\textbf{x}_i^s,\textbf{y}_i^s)\}_{i=1}^{|S|}$ from a source domain $\mathcal{D}^s$ and a set of \textit{unlabelled} samples $T=\{\textbf{x}_i^t\}_{i=1}^{|T|}$ from a target domain $\mathcal{D}^t$, where $\textbf{x}$ is a stance-bearing utterance and $\textbf{y}$ its stance label,
the goal is to learn a transferable (i.e., domain-invariant and stance-discriminative) dual-view stance feature $\textbf{f}_{\text{dual}}^t$ for any target sample $\textbf{x}^t$, and train a classifier $C_\text{stance}$ to predict its stance label $\textbf{y}^t$ with $\textbf{f}_{\text{dual}}^t$.

\section{DAN: Dual-View Adaptation Network}
\begin{figure}[h!]
	\centering
	\includegraphics[width=\columnwidth]{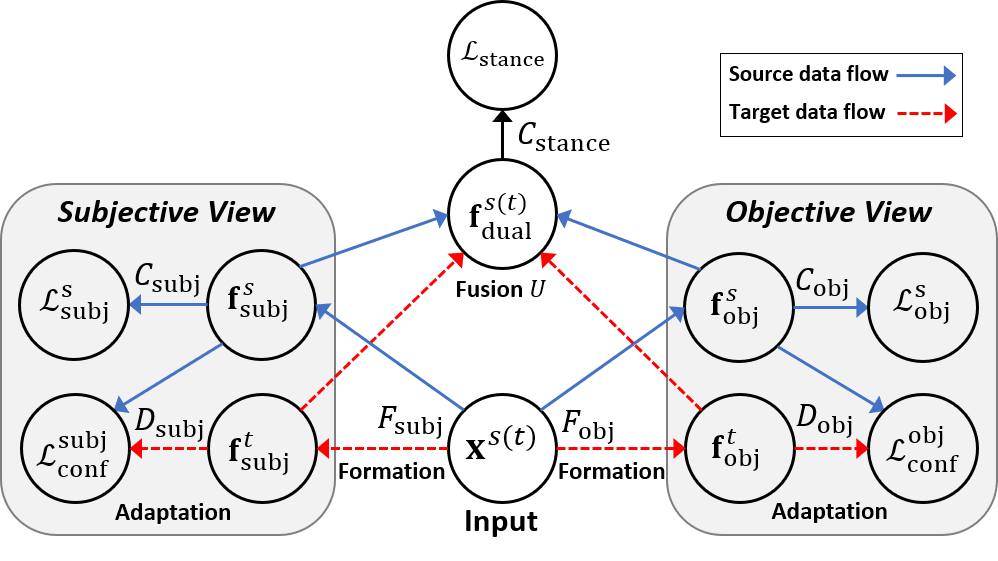}
	\vspace{-6mm}
	\caption{Model scheme of DAN. Both source and target utterances $\textbf{x}^{s(t)}$ are fed into the view functions $F_\text{subj}$ and $F_\text{obj}$ to produce the subjective and objective features $\textbf{f}_\text{subj}^{s(t)}$ and $\textbf{f}_\text{obj}^{s(t)}$, respectively. These features are then adapted separately and finally fused to form $\textbf{f}_\text{dual}^{s(t)}$ for the stance classification task.}
	\label{fig:dan}
	\vspace{-8mm}
\end{figure}

In this section, we introduce the dual-view adaptation network (DAN) for our task.
Figure~\ref{fig:dan} shows a sketch of DAN, which involves three view-centric subtasks designed for realising the above dual-view stance model, i.e., \textit{view formation}, \textit{adaptation}, and \textit{fusion}.
In \textit{view formation}, DAN learns the view functions $F_{\text{subj}}$ and $F_{\text{obj}}$ which create independent feature spaces for accommodating the subjective and objective features of input utterances, respectively.
In particular, it leverages \textit{multi-task learning} for obtaining the subjective/objective features that are also stance-discriminative.
Then, DAN performs \textit{view adaptation} to make each view feature invariant to the shift of domains.
This is done by solving a \textit{confusion maximisation} problem in each view which encourages the view features of both source and target samples to confuse the model about their origin domains (thus making the features domain-invariant).
Finally, DAN realises the transformation $U$ (i.e., Eq.~\ref{eq:dual_view_model}) in \textit{view fusion} that unifies both views to form the dual-view stance feature $\textbf{f}_\text{dual}$, which is used to produce the ultimate stance predictions.
Next, we elaborate each of these subtasks and show how they can be jointly trained to make the view features both stance-discriminative and domain-invariant.

\subsection{View Formation} The first and foremost subtask in DAN is to create the split subjective/objective views for the input utterances.
The challenge of this, however, lies in how to effectively extract the subjective and objective signals from an utterance $\textbf{x}$ which also reveal its stance polarity.
A straightforward solution is to utilise specialised lexica to identify words in $\textbf{x}$ that carry subjective and/or objective information.
For example, in \cite{mohammad2017stance}, several sentiment lexica were used to derive sentiment features for stance classification.
However, while there are many off-the-shelf lexica for finding the sentiment words, the ones for spotting the objective words/expressions are rarely available in practice.
Moreover, this approach does not guarantee finding stance-discriminative subjective/objective expressions.
Instead of searching the subjective/objective signals at the word level, we focus on learning the \textbf{stance-discriminative subjective and objective features} for the entire utterance.
In particular, we resort to \textit{multi-task learning} and formulate this view formation problem as learning stance-discriminative subjective/objective features with supervision from multiple related tasks.

Concretely, to learn the stance-discriminative subjective feature, we perform the stance classification task (main) together with a \textit{subjectivity classification} task (auxiliary), which predicts whether $\textbf{x}$ contains any subjective information.
Similarly, to learn the stance-discriminative objective feature, we perform the stance classification task (main) together with an \textit{objectivity classification} task (auxiliary), which predicts whether $\textbf{x}$ contains any objective information.
Formally, the above tasks can be expressed as follows,
\begin{equation}
\label{eq:sub_task}
\small
\begin{split}
\textit{\footnotesize View Learning:}\,\,\,&\textbf{f}_\text{subj}=F_\text{subj}(\textbf{x}; \theta_{F_\text{subj}}),\,\,\textbf{f}_\text{obj}=F_\text{obj}(\textbf{x}; \theta_{F_\text{obj}}) \\
\textit{\footnotesize Subj-Auxiliary Task:}\,\,\,&\hat{\textbf{y}}_\text{subj}=C_{\text{subj}}(\textbf{f}_\text{subj}; \theta_{C_\text{subj}}) \\
\textit{\footnotesize Obj-Auxiliary Task:}\,\,\,&\hat{\textbf{y}}_\text{obj}=C_{\text{obj}}(\textbf{f}_\text{obj}; \theta_{C_\text{obj}}) \\
\textit{\footnotesize Main Task:}\,\,\,&\hat{\textbf{y}}_\text{stance}=C_{\text{stance}}(\textbf{f}_\text{dual}; \theta_{C_\text{stance}})
\end{split}
\end{equation}
where a view function $F$ maps the input $\textbf{x}$ into its $d$-dimensional view feature $\textbf{f}$ with parameter $\theta_{F}$; $C$ denotes a classifier parameterised by $\theta_{C}$, and $\hat{\textbf{y}}$ a prediction.
To jointly learn these tasks, we minimise the negative log-likelihood of the ground truth class for each \textit{source} sample (as target samples are assumed to be unlabelled),
\begin{equation}
\label{eq:subj}
\small
\begin{split}
&\mathcal{L}_\text{stance}+\alpha\mathcal{L}_\text{subj}+\beta\mathcal{L}_\text{obj}=\\
&-\sum_{i=1}^{|S|}\textbf{y}_\text{stance}^{(i)}\ln\hat{\textbf{y}}_\text{stance}^{(i)} - \alpha\sum_{i=1}^{|S|}\textbf{y}_\text{subj}^{(i)} \ln\hat{\textbf{y}}_\text{subj}^{(i)} - \beta\sum_{i=1}^{|S|}\textbf{y}_\text{obj}^{(i)} \ln\hat{\textbf{y}}_\text{obj}^{(i)}
\end{split}
\end{equation}
where $\textbf{y}$s denote the true classes, and $\alpha$, $\beta$ the balancing coefficients.
Notice that both tasks $C_{\text{subj}}$ and $C_{\text{stance}}$ ($C_{\text{obj}}$ and $C_{\text{stance}}$) share the same underlying feature $\textbf{f}_{\text{subj}}$ ($\textbf{f}_{\text{obj}}$) for making predictions, and that $\textbf{f}_\text{dual}$ is a function of both $\textbf{f}_\text{subj}$ and $\textbf{f}_\text{obj}$ (Eq.\ref{eq:dual_view_model}); minimising Eq.\ref{eq:subj} thus encourages $\textbf{f}_{\text{subj}}$ ($\textbf{f}_{\text{obj}}$) to be stance-discriminative.

The ground-truth subjectivity and objectivity labels $\textbf{y}_\text{subj}$s and $\textbf{y}_\text{obj}$s are essential for computing the losses $\mathcal{L}_\text{subj}$ and $\mathcal{L}_\text{obj}$, respectively.
One can obtain gold standard labels from human annotations, which, however, is often a costly process.
To seek a cost-effective solution, we pre-train a subjective and an objective classifier with a publicly available subjectivity vs. objectivity corpus, and then use the pre-trained models to assign a subjective and an objectivity label to each utterance in our data as the \textit{silver} standard labels.
The benefits of this practice are two-fold.
First, it automates the label acquisition process;
Second, although these silver labels may be less reliable than the human-annotated gold standard ones, we find that the silver labels produced by the pre-trained models trained with large amounts of subjectivity/objectivity data are adequately informative in practice to indicate useful subjective/objective signals.
More details on obtaining such silver labels are discussed later in the experiments.


\subsection{View Adaptation} With both view features learned, we then perform feature alignment to match the distributions of source and target features $\{\textbf{f}_i^s\}$ and $\{\textbf{f}_i^t\}$ in each view, so that they both become invariant to the domain shift.
To achieve this, we introduce a \textit{confusion} loss, which \textit{adversarially} encourages the model (\textit{domain examiner} in particular) to be confused with the origin of a sample, i.e., which domain it comes from.
Then, by \textit{maximising} the confusion loss, the source and target features $\textbf{f}^s$ and $\textbf{f}^t$ would become asymptotically similar so as to confuse the model. 
The more alike $\textbf{f}^s$ and $\textbf{f}^t$ are, the more likely that the stance classifier $C_\text{stance}$ trained on $\textbf{f}^s$ would perform similarly well on $\textbf{f}^t$.
In this work, we experimented with two implementations of the confusion loss, both of which assess the level of confusion by measuring the discrepancy between domains in different ways.

The first one measures the $\mathcal{H}$-divergence between the domains, approximated as the classification loss incurred by the domain examiner to distinguish the source/target samples~\cite{ganin2015unsupervised}.
Specifically, the domain examiner learns a function $D^\mathcal{H}$, with parameter $\theta_{D^\mathcal{H}}$, that maps a feature $\textbf{f}=F(\textbf{x})$ to a binary class label indicating its domain.
Then, by maximising the following binary domain prediction loss with respect to $\theta_{D^\mathcal{H}}$, while minimising it with respect to $\theta_{F}$, one obtains the domain-invariant $\textbf{f}$,
\begin{equation}
\mathcal{L}_\text{conf}^\mathcal{H}=\sum_{i=1}^{|S|+|T|}\mathbbm{1}_{[\textbf{x}_i\in S]}\ln D^\mathcal{H}(\textbf{f}_i)+\mathbbm{1}_{[\textbf{x}_i\in T]}\ln(1-D^\mathcal{H}(\textbf{f}_i))
\end{equation}

The other one measures the Wasserstein distance between domains for the purpose of stabilising adversarial training~\cite{shen2018wasserstein}.
Specifically, the domain examiner learns a function $D^\mathcal{W}$, with parameter $\theta_{{D}^\mathcal{W}}$, that maps $\textbf{f}=F(\textbf{x})$ to a real number.
Then one can approximate the empirical Wasserstein distance by maximising the following domain critic loss with respect to $\theta_{{D}^\mathcal{W}}$,
\begin{equation}
\mathcal{L}_\text{conf}^\mathcal{W}=\frac{1}{|S|}\sum_{\textbf{x}_i\in S}D^\mathcal{W}(\textbf{f}_i)-\frac{1}{|T|}\sum_{\textbf{x}_i\in T}D^\mathcal{W}(\textbf{f}_i)
\end{equation}
To obtain the domain-invariant features, one minimises $\mathcal{L}_\text{conf}^\mathcal{W}$ with respect to $\theta_F$ when the domain examiner is trained to optimality~\cite{shen2018wasserstein}.

We create a different domain examiner for each of the subjective and objective views to achieve separate adaptation in each view.

\subsection{View Fusion} Finally, we combine the two adapted view features $\textbf{f}_\text{subj}$ and $\textbf{f}_\text{obj}$ to produce the dual-view stance feature $\textbf{f}_\text{dual}$.
The key is to select the best aligned dimensions of $\textbf{f}_\text{subj}$ and $\textbf{f}_\text{obj}$ to attain optimal combination.
For this, we merge $\textbf{f}_\text{subj}$ and $\textbf{f}_\text{obj}$ by learning a \textit{fusion function} $U(\cdot;\theta_{U})$, which serves as a realisation of the transformation $U$ in Eq.~\ref{eq:dual_view_model}, to weigh each dimension of $\textbf{f}_\text{subj}$ and $\textbf{f}_\text{obj}$ in two steps:
1) It learns a fusion vector $\textbf{g}$ via a feed-forward network with sigmoid activation: $\textbf{g}=\text{sigmoid}(\textbf{W}_u[\textbf{f}_\text{subj};\textbf{f}_\text{obj}]+\textbf{b}_u)$, where $[;]$ denotes vector concatenation, and $\theta_{U}=\{\textbf{W}_u,\textbf{b}_u\}$ the trainable parameters;
2) It merges the views using $\textbf{g}$ to deliver the dual-view stance feature $\textbf{f}_\text{dual}$,
\begin{equation}
\label{eq:fuse}
\textbf{f}_\text{dual}=\textbf{g}\odot\textbf{f}_\text{subj}+(\textbf{1}-\textbf{g})\odot\textbf{f}_\text{obj}
\end{equation}
 where $\odot$ is the element-wise product.
Note that during training, the fusion is applied to the source data only, as the target domain is assumed to be unlabelled.
After fusion, the dual-view stance feature $\textbf{f}_\text{dual}$ is used by the stance classifier $C_\text{stance}$ to produce a predicted stance label $\hat{\textbf{y}}_\text{stance}$ for the stance classification task (Eq.\ref{eq:sub_task}).

\subsection{Training and Prediction} To train DAN, we jointly optimise all the losses introduced above by using both the labelled source and unlabelled target data, so that the view features are learned to be both domain-invariant and stance-discriminative.
In this process, we need to minimise the classification losses ($\mathcal{L}_\text{stance}$, $\mathcal{L}_\text{subj}$, $\mathcal{L}_\text{obj}$) with respect to the classifiers ($C_\text{stance}$, $C_\text{subj}$, $C_\text{obj}$), the fusion function $U$, and the view functions ($F_\text{subj}$, $F_\text{obj}$) for obtaining stance-discriminative features, while \textit{adversarially} maximising the confusion losses ($\mathcal{L}_\text{conf}^\text{subj}$, $\mathcal{L}_\text{conf}^\text{obj}$) with respect to the domain examiners ($D_\text{subj}$, $D_\text{obj}$) and view functions ($F_\text{subj}$, $F_\text{obj}$) to make those features domain-invariant.
We thus formulate the training as a min-max game between the above two groups of losses, which involves alternating the following min and max steps until convergence:

\textbf{Min step}: Update the parameters of the view functions $\{\theta_{F_\text{subj}},\theta_{F_\text{obj}}\}$, the fusion function $\theta_U$, and the classifiers $\{\theta_{C_\text{stance}},\theta_{C_\text{subj}},\theta_{C_\text{obj}}\}$ with the following minimisation task,
\begin{equation}
\label{eq:overall_min}
\min_{\substack{\theta_{F_\text{subj}},\theta_{F_\text{obj}},\theta_U \\ \theta_{C_\text{stance}},\theta_{C_\text{subj}},\theta_{C_\text{obj}}}}\mathcal{L}_{\text{stance}}+\alpha\mathcal{L}_{\text{subj}}+\beta\mathcal{L}_\text{obj}+\gamma(\mathcal{L}_\text{conf}^\text{subj}+\mathcal{L}_\text{conf}^\text{obj})
\end{equation}
where $\alpha$, $\beta$, $\gamma$ are the balancing coefficients. 

\textbf{Max step}: Train the domain examiners $\{D_\text{subj},D_\text{obj}\}$ (could be $D^\mathcal{H}$ or $D^\mathcal{W}$) to optimality by maximising the confusion losses,
\begin{equation}
\label{eq:overall_max}
\max_{\theta_{D_\text{subj}},\theta_{D_\text{obj}}}\mathcal{L}_\text{conf}^\text{subj}+\mathcal{L}_\text{conf}^\text{obj}
\end{equation}

The above training process can be implemented with the standard back-propagation, the algorithm for which is summarised in Algorithm~\ref{algo:DAN_train}.
Once all parameters converge, the view feature $\textbf{f}_\text{subj}$ ($\textbf{f}_\text{obj}$) would become both domain-invariant and stance-discriminative, as the view function $F_\text{subj}$ ($F_\text{obj}$) has received gradients from both the confusion loss $\mathcal{L}_\text{conf}^\text{subj}$ ($\mathcal{L}_\text{conf}^\text{obj}$) and stance classification loss $\mathcal{L}_\text{stance}$ during back-propagation (lines 15$\sim$16 in Algorithm~\ref{algo:DAN_train}).
\begin{algorithm}[h!]
	\DontPrintSemicolon
	\footnotesize
	\SetKwRepeat{Do}{do}{while}
	\KwIn{source data $S$; target data $T$; batch size $m$; domain examiner training step $n$;  balancing coefficient $\alpha,\beta,\gamma$; learning rate $\lambda_1,\lambda_2$}
	\KwOut{$\theta_{F_\text{subj}},\theta_{F_\text{obj}},\theta_{D_\text{subj}},\theta_{D_\text{obj}},\theta_{C_\text{stance}},\theta_{C_\text{subj}},\theta_{C_\text{obj}},\theta_U$}
	Initialise the parameters of view functions, domain examiners, classifiers, and fusion function with random weights \; 
	\Repeat{$\theta_{F_\text{subj}},\theta_{F_\text{obj}},\theta_{D_\text{subj}},\theta_{D_\text{obj}},\theta_{C_\text{stance}},\theta_{C_\text{subj}},\theta_{C_\text{obj}},\theta_U$ converge}{
		Sample batch $\{\textbf{x}_i^s,\textbf{y}_i^s\}_{i=1}^m$, $\{\textbf{x}_i^t\}_{i=1}^m$ from $S$ and $T$\;
		\For{$k=1,...,n$} {
			\tcp{Maximisation Step}
			$\textbf{f}_\text{subj}^s=F_\text{subj}(\textbf{x}^s)$, $\textbf{f}_\text{obj}^s=F_\text{obj}(\textbf{x}^s)$\;
			$\textbf{f}_\text{subj}^t=F_\text{subj}(\textbf{x}^t)$, $\textbf{f}_\text{obj}^t=F_\text{obj}(\textbf{x}^t)$\;
			$\theta_{D_\text{subj}}\mathrel{{+}{=}}\lambda_1\nabla_{\theta_{D_\text{subj}}}\mathcal{L}_\text{conf}^\text{subj}(\textbf{f}_\text{subj}^s,\textbf{f}_\text{subj}^t)$\;
			$\theta_{D_\text{obj}}\mathrel{{+}{=}}\lambda_1\nabla_{\theta_{D_\text{obj}}}\mathcal{L}_\text{conf}^\text{obj}(\textbf{f}_\text{obj}^s,\textbf{f}_\text{obj}^t)$\;
		}
		\tcp{Minimisation Step}
		$\textbf{f}_\text{dual}^s \gets U(\textbf{f}_\text{subj}^s,\textbf{f}_\text{obj}^s)$\;
		$\theta_U\mathrel{{-}{=}}\lambda_2\nabla_{\theta_U}\mathcal{L}_\text{stance}(\textbf{f}_\text{dual}^s,\textbf{y}_\text{stance}^s)$\;
		$\theta_{C_\text{subj}}\mathrel{{-}{=}}\lambda_2\nabla_{\theta_{C_\text{subj}}}[\mathcal{L}_\text{stance}(\textbf{f}_\text{dual}^s,\textbf{y}_\text{stance}^s)+\alpha\mathcal{L}_\text{subj}(\textbf{f}_\text{subj}^s,\textbf{y}_\text{subj}^s)$]\;
		$\theta_{C_\text{obj}}\mathrel{{-}{=}}\lambda_2\nabla_{\theta_{C_\text{obj}}}[\mathcal{L}_\text{stance}(\textbf{f}_\text{dual}^s,\textbf{y}_\text{stance}^s)+\beta\mathcal{L}_\text{obj}(\textbf{f}_\text{obj}^s,\textbf{y}_\text{obj}^s)$]\;
		$\theta_{C_\text{stance}}\mathrel{{-}{=}}\lambda_2\nabla_{\theta_{C_\text{stance}}}\mathcal{L}_\text{stance}(\textbf{f}_\text{dual}^s,\textbf{y}_\text{stance}^s)$\;
		$\theta_{F_\text{subj}}\mathrel{{-}{=}}\lambda_2\nabla_{\theta_{F_\text{subj}}}[\mathcal{L}_\text{stance}(\textbf{f}_\text{dual}^s,\textbf{y}_\text{stance}^s)+\alpha\mathcal{L}_\text{subj}(\textbf{f}_\text{subj}^s,\textbf{y}_\text{subj}^s)+\gamma\mathcal{L}_\text{conf}^\text{subj}(\textbf{f}_\text{subj}^s,\textbf{f}_\text{subj}^t)]$\;
		$\theta_{F_\text{obj}}\mathrel{{-}{=}}\lambda_2\nabla_{\theta_{F_\text{obj}}}[\mathcal{L}_\text{stance}(\textbf{f}_\text{dual}^s,\textbf{y}_\text{stance}^s)+\beta\mathcal{L}_\text{obj}(\textbf{f}_\text{obj}^s,\textbf{y}_\text{obj}^s)+\gamma\mathcal{L}_\text{conf}^\text{obj}(\textbf{f}_\text{obj}^s,\textbf{f}_\text{obj}^t)]$\;
		
	}
	\caption{Adversarial Training of DAN}
	\label{algo:DAN_train}
\end{algorithm}

Once the training finishes, we are ready to make stance predictions on the target domain $\mathcal{D}^t$.
The prediction phase is more straightforward compared to the training, as it only involves chaining together the learned view functions $F_\text{subj}$ and $F_\text{obj}$, the fusion function $U$, and the stance classifier $C_\text{stance}$ to transform a target utterance $\textbf{x}^t\sim\mathcal{D}^t$ into a stance prediction: $\hat{\textbf{y}}^t=C_\text{stance}(U(F_\text{subj}(\textbf{x}^t),F_\text{obj}(\textbf{x}^t)))$.

\begin{table*}[h]
	\footnotesize
	\ra{1.2}
	\centering
	\caption{Performance (macro-$F1$ \%) of the baselines and their DAN-enhanced versions on various adaptation tasks. The highest performance on each task is underlined. The improvements of a DAN-enhanced method over its original version are shown in the parentheses, with the largest one on each task in bold.}
	\begin{tabular}{c|cccc|llll|c}
		\hline
		\textbf{S}\quad \textbf{T} 	  & \textbf{SO}    & \textbf{CORAL} &  \textbf{DANN}  & \textbf{WDGRL} & \multicolumn{1}{c}{\textbf{D-SO}} & \multicolumn{1}{c}{\textbf{D-CORAL}}  		       	& \multicolumn{1}{c}{\textbf{D-DANN}}   			      		& \multicolumn{1}{c|}{\textbf{D-WDGRL}} & \textbf{TO}\\
		\hline
		AT\quad CC & 31.96 & 48.66 &  49.45 & 48.39 &  35.43 (3.46)$^{**}$ & 54.97 (6.31)$^{***}$ 			  & \underline{55.78} (\textbf{6.32})$^{***}$& 53.36 (4.96)$^{***}$ & \multirow{4}{*}{59.18}\\
		FM\quad CC & 36.63 & 58.83 &  49.89 & 56.06 &  39.25 (2.61)$^{***}$ & \underline{62.04} (2.21)$^{***}$ & 52.25 (\textbf{2.35})$^{*}$ 			& 58.35 (2.28)$^{*}$ & \\
		HC\quad CC & 37.47 & 64.76 &  59.48 & 65.54 &  40.21 (2.74)$^{**}$ & \underline{71.84} (\textbf{7.07})$^{***}$ & 63.54 (4.06)$^{***}$ 			& 70.76 (5.21)$^{***}$ & \\
		LA\quad CC & 36.91 & 59.31 &  48.88 & 56.42 &  40.79 (3.88)$^{***}$ & \underline{62.17} (2.85)$^{*}$   & 53.51 (\textbf{4.62})$^{*}$ 			& 59.91 (3.48)$^{***}$ & \\
		\hline
		AT\quad HC & 35.11 & 63.12 &  66.90 & 66.08 &  38.05 (2.93)$^{*}$ & 67.87 (\textbf{4.75})$^{*}$ 		& \underline{70.05} (3.14)$^{*}$ 			& 69.34 (3.26)$^{**}$ & \multirow{4}{*}{68.33} \\
		CC\quad HC & 38.73 & 66.53 &  64.32 & 66.75 &  41.17 (2.44)$^{*}$ & 68.86 (2.32)$^{*}$ 				& 67.53 (\textbf{3.21})$^{***}$			& \underline{69.49} (2.73)$^{***}$ &  \\
		FM\quad HC & 44.65 & 68.10 &  69.75 & 75.08 &  46.97 (2.32)$^{*}$ & 73.12 (5.02)$^{**}$ 				& 76.77 (\textbf{7.01})$^{***}$			& \underline{77.72} (2.64)$^{**}$ & \\
		LA\quad HC & 38.05 & 59.21 &  63.31 & 55.61 &  40.64 (2.59)$^{*}$ & 63.78 (\textbf{4.57})$^{***}$ 	& \underline{66.90} (3.59)$^{**}$			& 59.33 (3.71)$^{**}$ & \\
		\hline
		AT\quad LA & 35.58 & 71.13 &  69.84 & 75.36 &  38.14 (2.56)$^{***}$ & 75.96 (4.82)$^{***}$ 			& 77.58 (\textbf{7.73})$^{***}$			& \underline{78.25} (2.89)$^{***}$ & \multirow{4}{*}{68.41}\\
		CC\quad LA & 42.47 & 62.75 &  74.04 & 69.26 &  45.27 (2.80)$^{*}$ & 68.99 (\textbf{6.23})$^{**}$		& \underline{76.60} (2.55)$^{***}$		& 72.88 (3.61)$^{***}$ & \\
		FM\quad LA & 43.15 & 68.60 &  67.47 & 69.37 &  46.14 (2.99)$^{***}$ & \underline{73.75} (\textbf{5.15})$^{***}$& 72.41 (4.93)$^{***}$			& 73.41 (4.03)$^{***}$ & \\
		HC\quad LA & 40.16 & 52.11 &  53.42 & 71.16 &  43.25 (3.09)$^{*}$ & 61.70 (\textbf{9.59})$^{***}$	& 61.05 (7.62)$^{**}$ 					& \underline{74.93} (3.77)$^{**}$ & \\
		\hline
		AT\quad FM & 34.37 & 65.10 &  52.77 & 62.91 &  37.91 (3.53)$^{}$ & \underline{70.71} (\textbf{5.60})$^{*}$& 56.43 (3.65)$^{*}$					& 66.22 (3.31)$^{*}$ & \multirow{4}{*}{61.49}\\
		CC\quad FM & 40.57 & 66.42 &  60.17 & 52.23 &  44.18 (3.60)$^{***}$ & \underline{69.29} (2.87)$^{***}$	& 65.33 (5.15)$^{***}$				& 58.08 (\textbf{5.85})$^{***}$ & \\
		HC\quad FM & 41.82 & 72.47 &  63.02 & 66.96 &  45.73 (3.90)$^{***}$ & \underline{74.59} (2.12)$^{***}$	& 71.77 (\textbf{8.74})$^{***}$		& 69.74 (2.78)$^{***}$ & \\
		LA\quad FM & 42.71 & 59.92 &  55.80 & 57.36 &  45.51 (2.79)$^{**}$ & \underline{62.52} (2.60)$^{**}$		& 58.67 (\textbf{2.87})$^{**}$		& 60.12 (2.75)$^{**}$ & \\
		\hline
		CC\quad AT & 31.29 & 73.14 &  64.09 & 64.95 &  35.15 (3.85)$^{***}$ & \underline{75.34} (2.20)$^{***}$	& 69.43 (\textbf{5.34})$^{***}$		& 67.05 (2.09)$^{***}$ & \multirow{4}{*}{70.81}\\
		FM\quad AT & 32.19 & 70.08 &  70.70 & 77.46 &  37.32 (5.13)$^{***}$ & 73.91 (3.82)$^{***}$				& 76.13 (\textbf{5.42})$^{***}$		& \underline{80.05} (2.59)$^{***}$ & \\
		HC\quad AT & 34.87 & 76.31 &  72.27 & 67.28 &  38.21 (3.34)$^{***}$ & \underline{81.16} (\textbf{4.84})$^{***}$& 74.37 (2.10)$^{***}$			& 71.22 (3.93)$^{***}$ & \\
		LA\quad AT & 42.43 & 62.89 &  74.04 & 71.39 &  45.09 (2.66)$^{**}$ & 69.37 (\textbf{6.48})$^{***}$		& \underline{79.44} (5.39)$^{***}$	& 74.05 (2.65)$^{**}$ & \\
		\hline
		\hline
		Average        & 38.06 & 64.47 & 62.48  & 64.78 &  41.22 (3.15) & \underline{69.09} (4.62)   			& 67.27 (\textbf{4.78})				& 68.21 (3.42) & 65.64\\
		\hline
		\multicolumn{10}{l}{(Two-tailed t-test: $^{***}\ p<0.01$; $^{**}\ p<0.05$; $^{*}\ p<0.1$)}\\
	
	\end{tabular}
	\label{tb:domain_shift}
\end{table*}

\section{Experiments}
In this section, we evaluate the performance of DAN on a wide range of adaptation tasks.
We first conduct a quantitative study on the overall cross-domain classification performance of DAN on all the tasks.
Then, a series of qualitative experiments is performed to further examine the various properties of DAN.

\subsection{Experimental Setup}
\noindent\textbf{Dataset:} To evaluate DAN, we utilise the dataset publicised by SemEval-2016 Task 6\footnote{\url{http://alt.qcri.org/semeval2016/task6/}} on tweet stance classification, which has been widely used for benchmarking stance classifiers.
It contains stance-bearing tweets on five different domains/topics: Climate Change is a Real Concern (CC: 564), Feminist
Movement (FM: 949), Hillary Clinton (HC: 984), Legalisation of Abortion (LA: 933), and Atheism (AT: 733)\footnote{The number of tweets in each domain is shown in the parentheses.}.
Each tweet in the dataset is associated with one of the three stance labels: \textit{favour}, \textit{against}, and \textit{neutral}.
On these domains, we construct the complete set of adaptation tasks over all 20 (S)ource$\rightarrow$(T)arget domain pairs.
For each $S\rightarrow T$ pair, we use 90\% tweets from $S$ and all from $T$ (without labels) as the training data, the rest 10\% from $S$ as the validation data, and all labelled tweets from $T$ for testing.

\noindent\textbf{Baselines:}
We consider the following approaches as our baselines: 1) \textbf{SO}: a source-only stance classification model based on a Bidirectional LSTM~\cite{augenstein2016stance}; it is trained on the source data only, without using any adaptation;
2) \textbf{CORAL}~\cite{sun2016deep}: it performs correlation alignment for minimising domain discrepancy by aligning the second-order statistics of the source/target distributions;
3) \textbf{DANN}~\cite{ganin2016domain}: an adversarial adaptation network that approximates the $\mathcal{H}$-divergence between domains;
and 4) \textbf{WDGRL}~\cite{shen2018wasserstein}: an adversarial adaptation network that approximates the Wasserstein distance between domains.

All the above methods are single-view based and can be enhanced by DAN, i.e., they can be extended by specific components of DAN to learn the subjective and objective views, forming their respective dual-view variants:
5) \textbf{D-SO}: SO trained with the two view functions $\{F_\text{subj}, F_\text{obj}\}$ and classifiers $\{C_\text{subj}, C_\text{obj}\}$, and the fusion function $U$ for combining the two views;
6) \textbf{D-CORAL}: besides adding $\{F_\text{subj}, F_\text{obj}, C_\text{subj}, C_\text{obj}, U\}$ to the model, CORAL is also extended with two CORAL losses~\cite{sun2016deep} for the objective and subjective views, respectively;
7) \textbf{D-DANN}: in \textit{view adaptation}, two $\mathcal{L}_\text{conf}^\mathcal{H}$-focused domain examiners $\{D^\mathcal{H}_\text{subj}, D^\mathcal{H}_\text{obj}\}$ are used to align the source/target data in the respective views;
8) \textbf{D-WDGRL}: two $\mathcal{L}_\text{conf}^\mathcal{W}$-focused domain examiners $\{D^\mathcal{W}_\text{subj}, D^\mathcal{W}_\text{obj}\}$ are used for \textit{view adaptation}; and 9) \textbf{TO}~\cite{li-caragea-2019-multi}: we finally include as a reference the in-domain results from a state-of-the-art target-only method on the same dataset used here\footnote{The in-domain result on a target domain from \cite{li-caragea-2019-multi} is measured on the official test set of that domain, while in this work the whole dataset (both official training and test sets) is used for testing. The results are partially comparable due to the shared test set used in both cases.}.

\subsection{Implementation Details}
\noindent\textbf{Model:}
Each view function $F$ is implemented as a RNN-based encoder to convert an utterance $\textbf{x}$ into its feature $\textbf{f}$.
In this encoder, each word $x_j\in\textbf{x}$ is first embedded into a $d_e$-dimensional word vector $\textbf{w}_j=W[x_j]$, where $W\in\mathbb{R}^{d_e\times V}$ is the word embedding matrix, $V$ is the vocabulary size, and $W[x]$ represents the $x$-th column of $W$.
Then, a bi-directional LSTM with hidden size $d_h$ is used to encode the word sequence $\textbf{w}=\{w_j\}$ as $\textbf{h}_j=\text{BiLSTM}(\textbf{h}_{j-1}, \textbf{h}_{j+1}, \textbf{w}_j)$, where $\textbf{h}_j\in\mathbb{R}^{2d_h}$ is the output for the $j$th time step.
Finally, a linear mapping is used to project each $\textbf{h}_j$ back to dimension $d_h$: $\textbf{h}_j=\textbf{W}_l \textbf{h}_j+\textbf{b}_l$, with $\textbf{W}_l\in\mathbb{R}^{d_h\times 2d_h},\textbf{b}_l\in\mathbb{R}^{d_h}$ the trainable parameters.
Each of the classifiers $\{C_\text{subj}, C_\text{obj}, C_\text{stance}\}$ and domain examiners $\{D^\mathcal{H}_\text{subj}, D^\mathcal{H}_\text{obj},D^\mathcal{W}_\text{subj}, D^\mathcal{W}_\text{obj}\}$ is realised with a unique two-layer feed-forward network with hidden size $d_f$ and ReLU activation.

\noindent\textbf{Training:} 
The pre-trained GloVe word vectors ($d_e$=200, glove.twitter.27B) are used to initialise the word embeddings, which are fixed during training.
Batches of 8 samples from each domain are used in each training step.
All models are optimised using Adam~\cite{kingma2014adam}, with a varied learning rate based on the schedule: $lr=10^{-3}\cdot\min(\frac{1}{\sqrt{step}},\frac{step}{warmup})$.
$\alpha,\beta,\gamma$ are set equally to 0.1 for balancing the corresponding loss terms. 
Each compared method is ran ten times with random initialisations under different seeds.
The mean value of the evaluation metric is reported.
The hidden sizes of LSTM ($d_h$) and feed-forward network ($d_f$) are randomly sampled from the interval $[100, 300]$ upon each run.
A light dropout (0.1) is used.
We pre-train a subjectivity classifier and an objectivity one to obtain the silver standard subjectivity/objectivity labels for our data.
A widely-used subjectivity vs. objectivity dataset~\cite{pang2004sentimental} is used for the pre-training, which consists of 5000 subjective sentences (movie reviews) and 5000 objective sentences (plot summaries).
The pre-training is implemented with the FastText library~\cite{joulin2016bag}.

\subsection{Quantitative Results}

We report the overall classification performance of DAN and the baselines on all adaptation tasks in Table~\ref{tb:domain_shift}.

First, all the DAN-enhanced methods (D-X) are shown to improve over their original versions on all adaptation tasks, with the improvements ranging from 2.12\% to 9.59\% at different significant levels.
This shows that it is empirically superior to apply DAN for domain adaptation on stance data, and that the separate feature alignment for the subjective and objective stance expressions is effective in alleviating the domain drift issue.
Among these methods, D-CORAL obtains generally better performance than others on half of all tasks (10/20), suggesting that most domains of this dataset can be much aligned by the second-order statistics (feature covariances) of the samples.

Second, the improvements achieved by the adaptative models (D-CORAL, D-DANN, and D-WDGRL) are generally higher than those by the non-adaptative model (D-SO).
This suggests that the feature alignment between the source and target data (adaptation) can benefit more from the dual-view modelling of DAN than the source-only learning (no adaptation).
This points out the key benefit of DAN that the features could be more easily aligned in the split views.

Finally, we notice that the results on certain target domains (e.g., tasks with CC as the target domain) are generally worse than those on others.
One possible reason for this could be related to the inherent complexity of the data distributions of different domains; for example, the in-domain performance on CC is reportedly the poorest among all the five domains~\cite{mohammad2016semeval}, thus probably making it the most challenging to transfer knowledge from other domains to it.

\begin{figure}[!b]
	\begin{tabular}{c@{}}
		\small \rotatebox[origin=c]{90}{CC$\rightarrow$HC}
	\end{tabular}
	\begin{tabular}{@{}c@{}}
		\includegraphics[width=0.30\columnwidth]{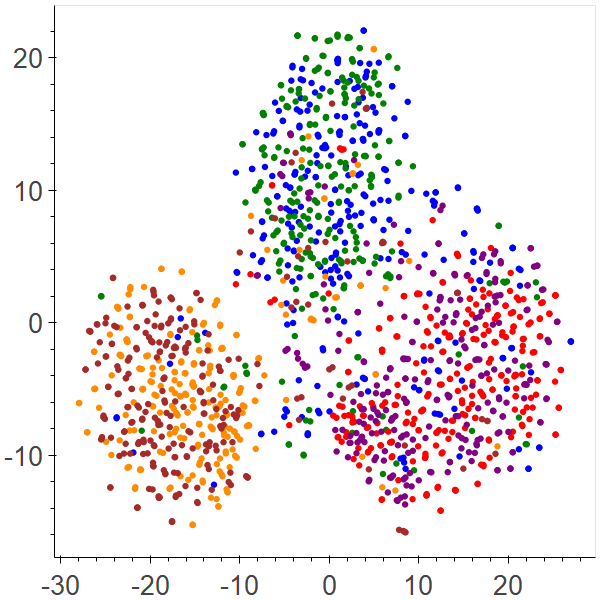}\\
		\scriptsize (a) CORAL
	\end{tabular}
	\begin{tabular}{@{}c@{}}
		\includegraphics[width=0.30\columnwidth]{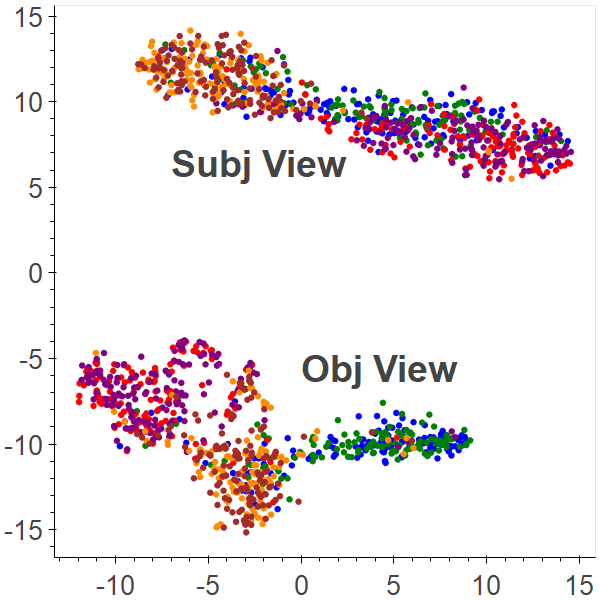}\\
		\scriptsize (b) D-CORAL (\textbf{split})
	\end{tabular}
	\begin{tabular}{@{}c@{}}
		\includegraphics[width=0.30\columnwidth]{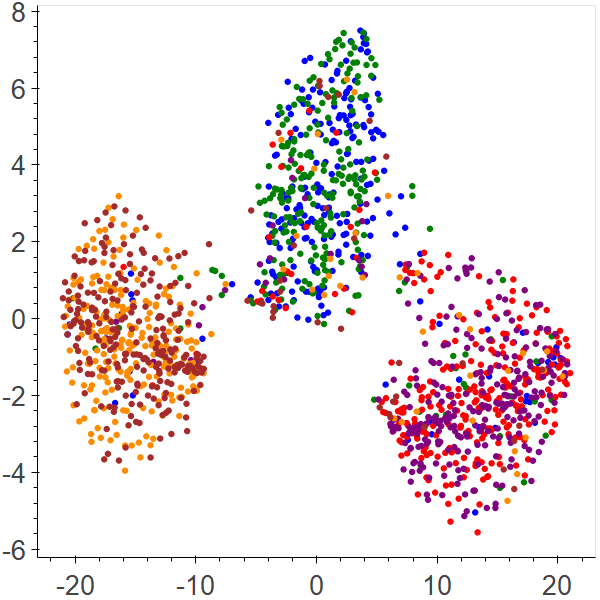}\\
		\scriptsize (c) D-CORAL (\textbf{dual})
	\end{tabular}

	\vspace{\floatsep}
	\begin{tabular}{c@{}}
		\small \rotatebox[origin=c]{90}{AT$\rightarrow$FM}
	\end{tabular}
	\begin{tabular}{@{}c@{}}
		\includegraphics[width=0.30\columnwidth]{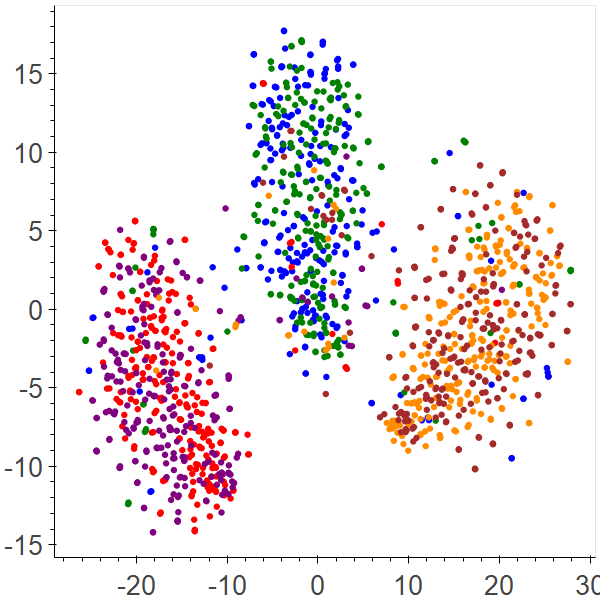}\\
		\scriptsize (d) DANN
	\end{tabular}
	\begin{tabular}{@{}c@{}}
		\includegraphics[width=0.30\columnwidth]{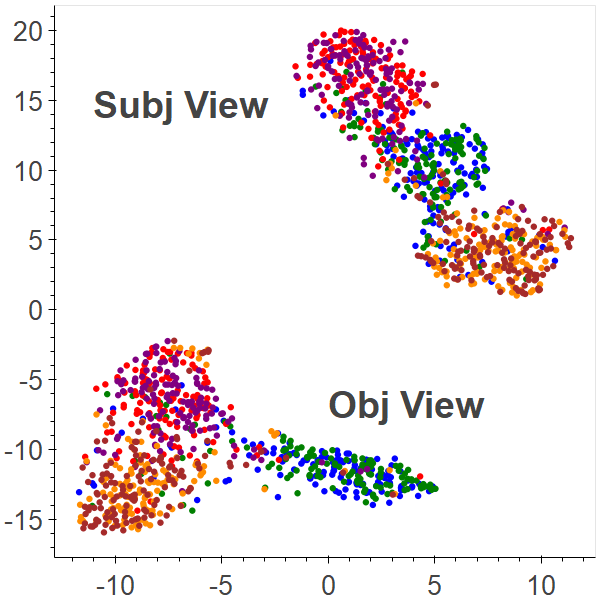}\\
		\scriptsize (e) D-DANN (\textbf{split})
	\end{tabular}
	\begin{tabular}{@{}c@{}}
		\includegraphics[width=0.30\columnwidth]{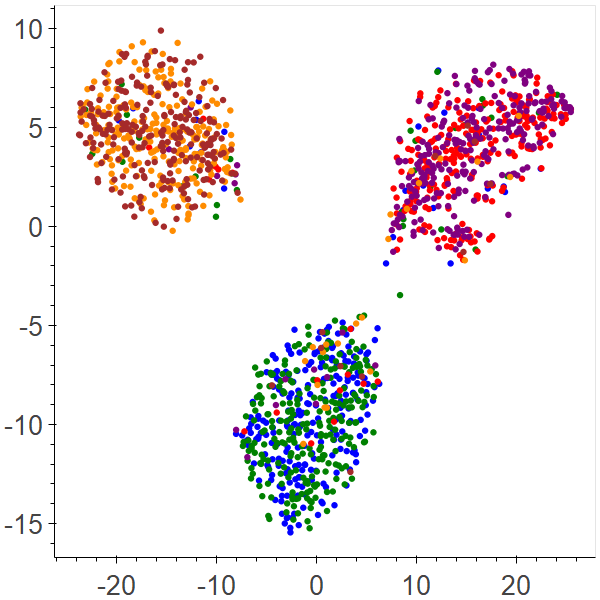}\\
		\scriptsize (f) D-DANN (\textbf{dual})
	\end{tabular}

	\vspace{\floatsep}
	\begin{tabular}{c@{}}
	\small \rotatebox[origin=c]{90}{HC$\rightarrow$LA}
	\end{tabular}
	\begin{tabular}{@{}c@{}}
		\includegraphics[width=0.30\columnwidth]{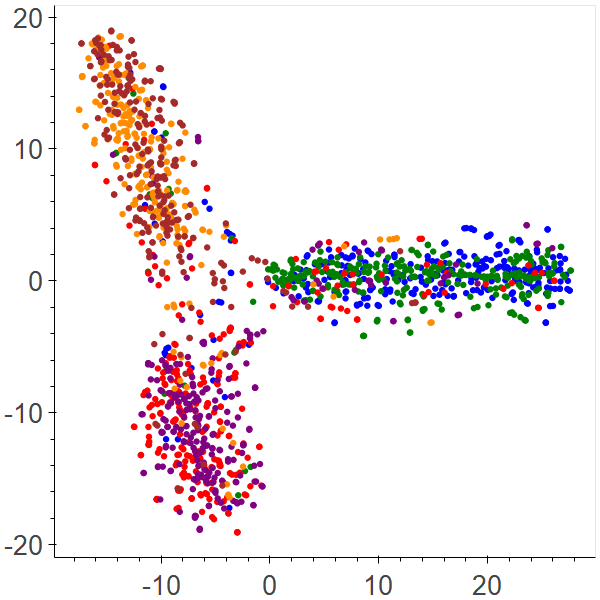}\\
		\scriptsize (g) WDGRL
	\end{tabular}
	\begin{tabular}{@{}c@{}}
		\includegraphics[width=0.30\columnwidth]{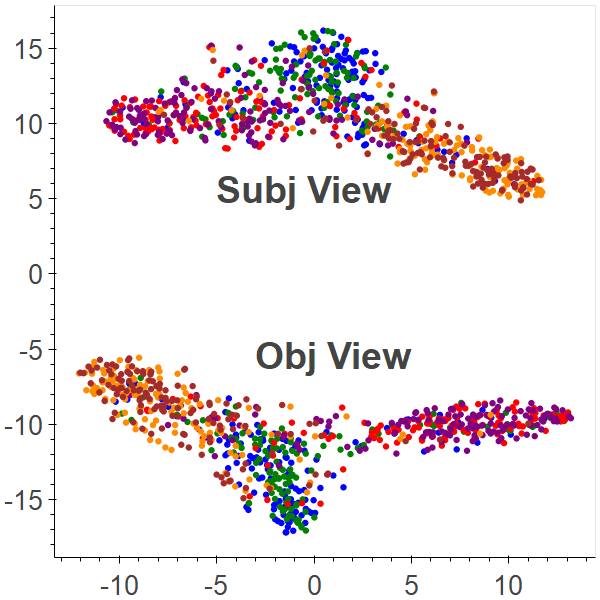}\\
		\scriptsize (h) D-WDGRL (\textbf{split})
	\end{tabular}
	\begin{tabular}{@{}c@{}}
		\includegraphics[width=0.30\columnwidth]{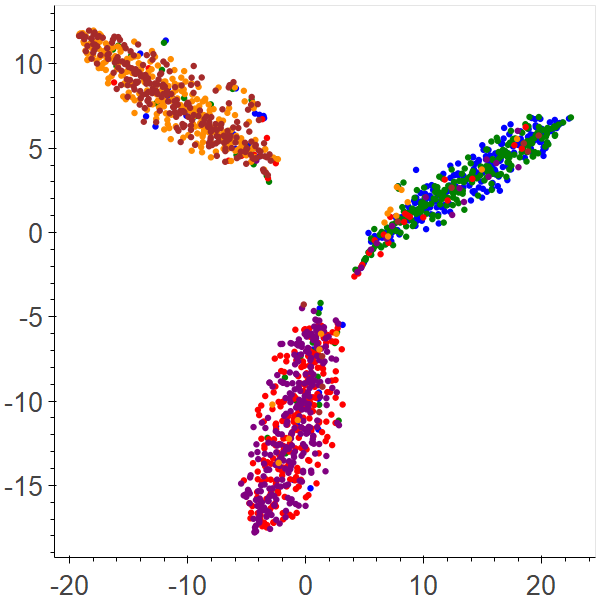}\\
		\scriptsize (i) D-WDGRL (\textbf{dual})
	\end{tabular}
	\captionof{figure}{The t-SNE plots of the feature distributions of the source and target samples learned by the three adaptative methods, $\{$CORAL, DANN, WDGRL$\}$, and their DAN-enhanced versions $\{$D-CORAL, D-DANN, D-WDGRL$\}$. 
		Source samples are coloured by orange (\textit{favour}), blue (\textit{against}), and red (\textit{neutral}), while target samples by brown (\textit{favour}), green (\textit{against}), and purple (\textit{neutral}).
		Each row compares the plots of a baseline method (X) and its DAN-enhanced version (D-X) on a randomly selected task.
		For each of the DAN-enhanced methods (D-X), we plot both the intermediate subjective/objective features $\{\textbf{f}_\text{subj},\textbf{f}_\text{subj}\}$ (Eq.~\ref{eq:sub_task}) in its \textbf{split} view and the ultimately fused dual-view stance features $\textbf{f}_\text{dual}$ (Eq.~\ref{eq:fuse}) in its \textbf{dual} view.}
	\label{tbl:feature_space}
\end{figure}


\subsection{Qualitative Analysis}
To gain a better understanding of what leads to DAN's superiority over the single-view methods, we conduct further experiments to qualitatively evaluate the properties of DAN.

\textbf{(1) Visualising view features in DAN's enhanced feature space.}
We first derive insights into what has been learned in the feature space of DAN that makes its view features more transferable.
For this, we plot the feature distributions of the source and target samples learned by all compared (adaptative) methods in Figure~\ref{tbl:feature_space}.

The first row shows the case of CORAL and D-CORAL, where we observe how DAN improves features' stance discriminating power.
First, the plot of CORAL (Figure~\ref{tbl:feature_space}.a) exhibits a good separation between each of the three stance classes except that between \textit{against} (blue/green) and \textit{neutral} (red/purple), as the boundary between these two is rather blurred by many source \textit{against} features (blue) invading the \textit{neutral} region.
The reason behind this seems to be revealed in the split view plot of D-CORAL (Figure~\ref{tbl:feature_space}.b), where a similar pattern (\textit{against} and \textit{neutral} classes overlap) is also seen in the \textbf{subj}ective view of D-CORAL.
Fortunately, its \textbf{obj}ective view remedies this issue, by yielding features that better separate the problematic classes (i.e., \textit{against} vs. \textit{neutral}).
As a result, the overall fused features in the dual view plot of D-CORAL (Figure~\ref{tbl:feature_space}.c) become more stance-discriminative, leading to a much better separation of all classes.

The second row demonstrates the case of how DAN improves features' domain-invariance with better feature alignment.
In this case, the features learned by DANN already shows good discrimination in stance classes (Figure~\ref{tbl:feature_space}.d), but the alignment of the source and target features seems less effective, as in each class they tend to scatter over a relatively large area, making the distance longer between the two feature distributions\footnote{Note that it happens to be the case that each stance class from one domain matches its counterpart in the other domain in this experiment (i.e., source \textit{favour} vs. target \textit{favour}, etc.). The feature alignment in DAN is class-agnostic; it does not assume any kind of match between particular class labels a priori, nor does it impose any such constraint during training.}.
In contrast, both the subjective and objective views of D-DANN produce more compact feature distributions within each class (Figure~\ref{tbl:feature_space}.e), suggesting smaller distances obtained between the source and target features.
Consequently, we observe a much stronger feature alignment achieved within each class in the ultimate dual view of D-DANN (Figure~\ref{tbl:feature_space}.f).

Finally, the last row shows a more successful case of DAN than before, where it improves both the domain invariance and stance-discriminative power of the source and target features.
As shown, although WDGRL already achieves superior feature learning than the previous single-view cases, D-WDGRL manages to produce even better dual-view features (Figure~\ref{tbl:feature_space}.i), which are shown to be more compact within the same classes and separable over different classes.

Overall, the above results suggest that, compared to the indiscriminate treatment applied in the single-view methods, the separate characterisation of the subjective and objective stance expressions in the split views of DAN could learn more transferable features for better adapting stance classifier across domains.

\textbf{(2) Ablation analysis: subjective view vs. objective view.}
As the two views of DAN have shown to sometimes learn features with distinct transferability in the previous experiment, it is necessary to further examine how differently each view contributes to the overall adaptation.
To this end, we build two ablation variants of DAN, namely D-DANN-SUBJ and D-DANN-OBJ (DANN is used here as an exemplified base for DAN), with each working in a specific view only.
Specifically, for D-DANN-SUBJ (D-DANN-OBJ), we keep the subjectivity (objectivity) classifier for learning the view feature and the subjectivity (objectivity) domain examiner for making the feature domain-invariant.
\begin{figure}[h!]
	\centering
	\includegraphics[width=\columnwidth]{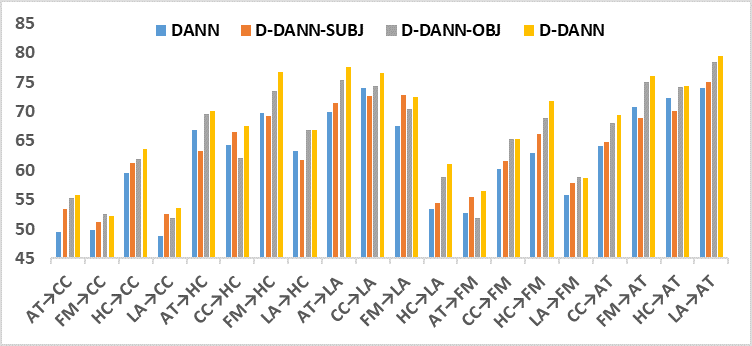}
	\vspace{-5mm}
	\caption{Performance of the ablated variants of DAN.}
	\label{fig:ablation}
	\vspace{-3mm}
\end{figure}
Figure~\ref{fig:ablation} shows the results of all ablated D-DANN variants over the 20 adaptation tasks.
As we can see, D-DANN-OBJ surpasses D-DANN-SUBJ on most of the tasks (16/20), with an averaged improvement of 3.38\% on its dominating tasks.
This indicates that the objective information in stance expressions is relatively more transferable than the subjective one.
Since much of the subjectivity expressed in this dataset appears to be sentiments/emotions, this finding is somewhat consistent with previous studies on stance and sentiment analysis~\cite{sobhani2016detecting,mohammad2017stance}, where the utility of sentiments to stance recognition is sometimes limited.
This may occur when the sentiments and stances of the same utterances do not correlate well~\cite{mohammad2017stance}, potentially causing the model to collapse in the case of disagreed subjectivity and stance signals. 
As a result,  the subjective features alone could sometimes degrade the overall performance (D-DANN-SUBJ underperforms DANN on 6/20 tasks).
Indeed, the objective information such as facts tend to be more stance-indicative, since the reasons (usually stating some facts) that support a stance are often found to be more specific to that stance~\cite{hasan2014you,xu2019recognising}.
Finally, we observe that D-DANN, the full model combining both views, gives the best performance on almost all tasks (18/20), suggesting that \textit{view fusion} in DAN is essential for obtaining the best of both views.

\textbf{(3) Reduced domain discrepancy.}
Achieving good feature alignment is the key to successful domain adaptation.
The visualisation in Figure~\ref{tbl:feature_space} already shows some evidence for the superiority of DAN in matching the source and target features.
Here we provide further analysis to quantify how good such feature matching is.
In particular, we use the Proxy $\mathcal{A}$-distance (PAD)~\cite{ben2007analysis} as the metric, which is a common measure for estimating the discrepancy between a pair of domains.
It is defined by $2(1-2\epsilon)$, where $\epsilon$ is the generalisation error on the problem of distinguishing between the source and target samples.
Following the same setup in \cite{ganin2015unsupervised}, we compute the PAD value by training and testing a linear SVM using a data set created from both source and target features of the training examples.
For the DAN-enhanced methods,  the fused dual-view features $\textbf{f}_\text{dual}$ are used.

\begin{figure}[t!]
	\centering
		\begin{tabular}{@{}c@{}}
			\includegraphics[width=0.47\columnwidth]{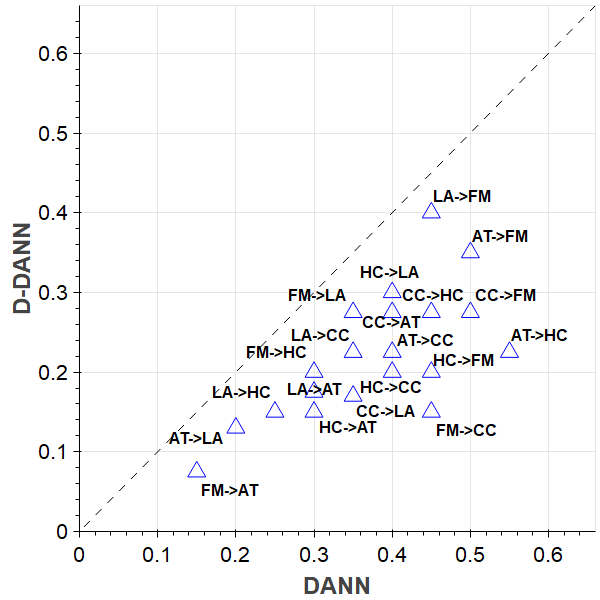}
		\end{tabular}
		\begin{tabular}{@{}c@{}}
			\includegraphics[width=0.47\columnwidth]{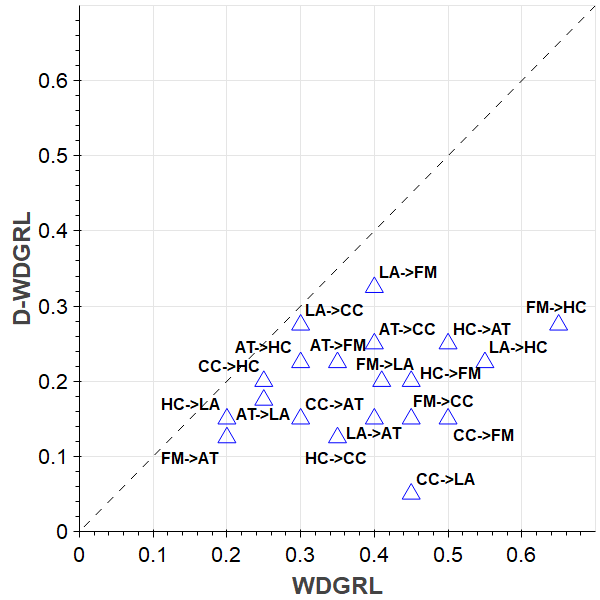}
		\end{tabular}
	\vspace{-3mm}
	\caption{Proxy $\mathcal{A}$-distance between every pair of the evaluated domains.}
	\label{fig:a_distance}
	\vspace{-3mm}
\end{figure}

Figure~\ref{fig:a_distance} displays the results of comparing the PAD values of  the two DAN-enhanced methods, D-DANN and D-WDGRL, with their original versions on the 20 adaptation tasks.
As shown, both D-DANN and D-WDGRL achieve lower PAD values than their original counterparts across all the tasks.
This validates the previous observations in Figure~\ref{tbl:feature_space} that the source/target dual-view features learned by D-DANN and D-WDGRL are better aligned than the respective cases of DANN and WDGRL.
Therefore, both the quantitative (Figure~\ref{fig:a_distance}) and qualitative (Figure~\ref{tbl:feature_space}) results manifest the benefit of the proposed dual-view adaptation in matching stance data of different domains.

\section{Related Work}
In-domain stance classification has been studied in the context of online debate forums~\cite{somasundaran2010recognizing,walker2012stance,hasan2013stance} and social media~\cite{augenstein2016stance,du2017stance,mohammad2017stance}. 
Recently, deep neural networks have been applied to learn rich representations of the stance-bearing utterances~\cite{augenstein2016stance,du2017stance,mohammad2017stance,sun2018stance}, which could be further enhanced by incorporating the subjective and objective supervision into the representation learning.

While domain adaption has been successful in the related task of cross-domain sentiment classification~\cite{pan2010cross,li2018hierarchical,peng2018cross}, its efficacy in the stance context is much less explored.
Along this line, pre-trained stance classifiers are fine-tuned for predicting stances on new domains~\cite{zarrella2016mitre}.
Attention-based models are built to extract salient information from the source data, which is expected to also work well on a related target domain~\cite{xu2018cross}.
In contrast, we take a different approach to exploiting existing knowledge, by making features invariant to the shift of domains.

Adversarial approaches have recently gained popularity in domain adaptation for aligning feature distributions~\cite{long2015learning,ganin2016domain,tzeng2017adversarial,shen2018wasserstein}.
In these approaches, a global feature space is solely induced to coarsely match the data from different domains.
By contrast, we explore the potential for finding a more fine-grained alignment of the domains by creating split feature spaces (views) to fully characterise the subjective and objective stance expressions.

The interaction between stance and subjectivity in stance classification has been studied recently~\cite{sobhani2016detecting,mohammad2017stance}, where the sentiment subjectivity shows its potential for predicting stances, although it is not as useful for stance classification as it is for sentiment classification.
There are few efforts on exploring the utility of objective information in stance expressions.
Some research examines reasons mentioned in the stance-bearing utterances~\cite{hasan2014you,boltuvzic2014back}, which often contain factual information for supporting the stances.
Different from all the above efforts, we leverage both subjective and objective information for better capturing the variety of stance expressions.

\section{Conclusions and Future Work}
In this paper, we propose the dual-view adaptation network, DAN, to adapt stance classifiers to new domains, which learns a subjective view and an objective view for every input utterance.
We show that DAN allows existing single-view methods to acquire the dual-view transferability, and that it empirically improves those methods on various adaptation tasks.
A series of qualitative analyses on the properties of DAN shows that more fine-grained adaptation for stance data can lead to more reduced domain discrepancies and finer stance discrimination, and that a proper view fusion is necessary for obtaining better overall features by leveraging the best of both views.

In the future, our work could be extended in several ways.
First, we plan to evaluate our method on more diverse stance datasets with different linguistic properties. 
For instance, the utterances in posts of online debate forums~\cite{somasundaran2010recognizing} are typically longer, which may pose 
new challenges such as capturing dependencies across multiple sentences as well as richer subjective/objective expressions.
Second, as DAN is input-agnostic (as long as the input is feature vectors), it would be interesting to apply it to other scenarios suitable for dual-view modelling.
One example is modelling the social network user behaviours~\cite{benevenuto2009characterizing} where the networks of users together with their interactions provide a dual-view of their behaviours.
Finally, it is possible that DAN could be extended for multi-view adaptation, e.g., by adding more view functions and domain examiners.

\ack We would like to thank all anonymous referees for their constructive comments. We would also like to thank Xiang Dai for his helpful comments on drafts of this paper.

\bibliography{ecai}
\end{document}